\documentclass[a4paper,fleqn]{cas-dc}

\usepackage[numbers]{natbib}
\usepackage{float}
\usepackage{multirow}
\usepackage[table]{xcolor}
\usepackage{array}

\definecolor{lightblue}{RGB}{220,235,247}

\renewcommand{\floatpagefraction}{0.8}

\hypersetup{
  colorlinks=true,
  linkcolor=red,
  citecolor=blue,
  urlcolor=blue
}
\def\tsc#1{\csdef{#1}{\textsc{\lowercase{#1}}\xspace}}
\tsc{WGM}
\tsc{QE}

\begin{document}
\setlength{\mathindent}{0pt}
\let\WriteBookmarks\relax
\def\floatpagefraction{1}
\def\textpagefraction{.001}
\let\printorcid\relax

\shorttitle{ }    

\shortauthors{K. Wang et al.}

\title[mode = title]{SDGIC: A Semantic Disambiguation-Guided Generative Image Compression Method for Ultra-Low Bitrates}

\tnotemark[1]

\tnotetext[1]{This work was supported in part by the National Natural Science Foundation of China under Grant 62471376.}

\author[1]{Kaile Wang}
\ead{kaile.wang@stu.xjtu.edu.cn}

\author[1]{Lijun He}

\author[2]{Haisheng Fu}

\author[1]{Haixia Bi}

\author[1]{Fan Li}
\cormark[1]
\ead{lifan@mail.xjtu.edu.cn}

\address[1]{School of Information and Communications Engineering,
Xi'an Jiaotong University, Xi'an 710049, China}

\address[2]{Department of Electrical and Computer Engineering,
Faculty of Applied Science, The University of British Columbia,
Vancouver, BC, Canada}

\cortext[1]{Corresponding author.}

\begin{abstract}
Generative image compression has recently shown impressive perceptual quality, but often suffers from semantic inconsistency at ultra-low bitrates (bpp < 0.05), limiting its reliable deployment in bandwidth-constrained scenarios such as 6G semantic communications. This inconsistency stems from incomplete guidance information, which introduces semantic ambiguity into the generation process and may lead to natural-looking but source-inconsistent content. In this work, we propose a Semantic-Disambiguation-Guided Generative Image Compression (SDGIC) framework to constrain diffusion-based reconstruction at ultra-low bitrates. Specifically, SDGIC compresses the source image into three compact and complementary guidance streams: a concise text caption for global semantics, a highly compressed image (HCI) for dense visual evidence, and Reconstruction-Aware Semantic Residual Tokens (RSRTs) for reconstruction-relevant residual semantics that remain ambiguous under the text caption and HCI conditions. The RSRTs are directly optimized toward the downstream denoising objective, enabling them to provide source-specific semantic constraints for disambiguating diffusion-based reconstruction. To inject these three guidance streams into the generation process effectively, we design a Dual-Path Conditioned Diffusion Decoder (DPCD), which uses cross-attention for semantic conditions and ControlNet residuals for dense visual guidance. Extensive experiments demonstrate that SDGIC improves semantic consistency at ultra-low bitrates while maintaining favorable perceptual quality (e.g., AFINE$\downarrow$ by 23.4\% on the CLIC2020 dataset).
\end{abstract}


\maketitle

\section{Introduction}
\begin{figure*}[t]
    \centering
    \includegraphics[width=\textwidth]{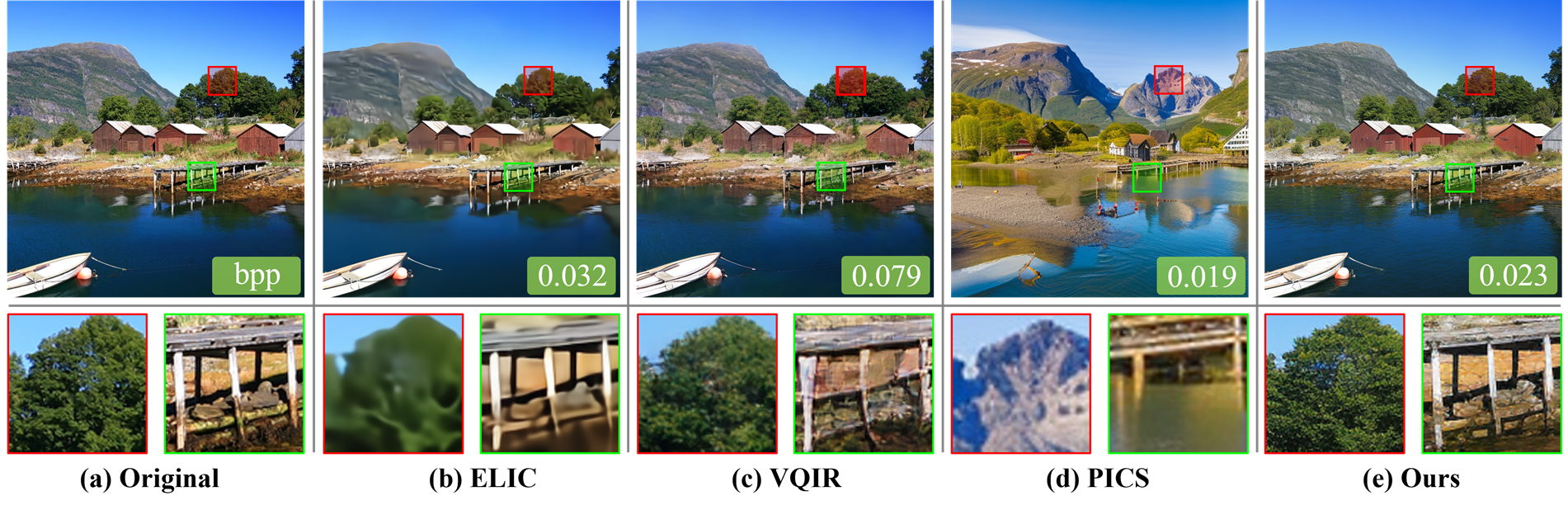}
    \caption{Visual comparison of different methods at ultra-low bitrates. (a) presents the original image, (b) ELIC\cite{he2022elic} suffers from blurring and oversmoothing, (c) VQIR\cite{wei2024toward} exhibits artifacts, (d) PICS\cite{lei2023text+} exhibit semantic deviations, while (e) our method achieves superior perceptual quality and semantic consistency.}
    \label{fig:fig1}
\end{figure*}

The growth of bandwidth-constrained visual applications, such as deep-space exploration\cite{xu2020optimal}, satellite Internet\cite{routray2020lossless}, and emerging 6G semantic communications\cite{shafi2026semantic}, has created a growing demand for ultra-low-bitrate image compression. In these scenarios, reconstructions must preserve high perceptual quality\cite{zhou2025perceptual} and semantic consistency\cite{raman2020compressnet,santurkar2018generative,chen2024generative} with the source image under stringent bit budgets. Conventional\cite{wallace1991jpeg,bellard2015bpg} and VAE-based codecs\cite{Lu2024preprocessing,li2025learned,bao2023taylor,fu2023learned} are primarily optimized to minimize pixel-level distortion and tend to discard high-frequency information that cannot be accurately encoded at ultra-low bitrates. This results in severe blurring and structural degradation, as illustrated in Fig.\,\ref{fig:fig1}(b). Consequently, achieving image reconstruction with both high perceptual quality and semantic consistency at ultra-low bitrates remains a significant challenge\cite{li2024misc}.

Generative image compression offers a promising solution to this challenge\cite{lei2023text+,li2026structure,grassucci2023generative}. Unlike conventional compression methods that rely on transmitted bits to recover image content, generative compression exploits priors learned through large-scale pretraining and requires only compact image-specific guidance to be transmitted. Conditioned on this guidance, the decoder can synthesize high-frequency details that are not explicitly encoded, thereby enabling perceptually natural image reconstruction at ultra-low bitrates. 
However, this advantage also introduces a critical risk: ultra-compact guidance often leaves the reconstruction underconstrained, so multiple visually plausible outputs may satisfy the same transmitted condition. We refer to this phenomenon as semantic ambiguity, where the decoder may rely excessively on learned priors and generate source-inconsistent content.

Early generative image compression relied on Generative Adversarial Networks (GANs)\cite{goodfellow2014generative}, which suffered from training instability and tended to synthesize statistically plausible but semantically inconsistent artifacts at ultra-low bitrate\cite{muckley2023improving,wei2024toward}, as shown in Fig.\,\ref{fig:fig1}(c). 
Existing diffusion-based image compression methods\cite{yang2023lossy,gao2024rate} typically employ text\cite{lei2023text+}, sketches, edge maps\cite{grassucci2023generative}, or heavily quantized latent\cite{careil2023towards} representations, etc., to guide image reconstruction. However, each type of guidance captures only a limited subset of the source image, leaving the reconstruction underconstrained. For example, text provides global semantics, such as scenes and objects, but cannot accurately capture color distributions or spatial layouts. Sketches and edge maps preserve sparse structures while discarding textures and dense appearance cues. Heavily quantized latent representations become blocky at ultra-low bitrates, leaving many critical local structures and instance-specific details unavailable to the generative model. As a result, semantic ambiguity remains unresolved in diffusion-based reconstruction, resulting in the generation of source-inconsistent content, as shown in Fig.\,\ref{fig:fig1}(d). Consequently, how to adequately constrain the generation process to reduce semantic ambiguity and preserve semantic consistency at ultra-low bitrates remains a critical challenge\cite{careil2023towards}.

Therefore, we propose Semantic-Disambiguation-Guided Generative Image Compression (SDGIC). At the encoder, the source image is compressed into three compact and complementary guidance, including a text caption, a highly compressed image (HCI), and Reconstruction-Aware Semantic Residual Tokens (RSRTs). At the decoder, a Dual-Path Conditioned Diffusion Decoder (DPCD) integrates these guidance streams to jointly constrain diffusion denoising.
Specifically, SDGIC constructs three information streams according to their distinct roles in resolving ambiguity during generation: 1) The concise text caption serves as a Global Semantic Anchor, providing the main objects, scene, and contextual relationships at a low bitrate cost. 2) The HCI serves as a Dense Visual Anchor, preserving visual evidence such as color, spatial layout, structure, and low-level appearance through rate-distortion optimization. 3) We further introduce RSRTs, generated by a Reconstruction-Aware Semantic Tokenizer (RAST) from the continuous visual representation of the source image. RSRTs target the residual ambiguities that remain after conditioning on the caption and HCI. Here, ``residual'' denotes their conditional role rather than explicit feature subtraction: they encode reconstruction-relevant cues not determined by the other two streams. Optimized through the downstream denoising objective, RSRTs learn source-specific cues that help select the source-consistent reconstruction from multiple plausible candidates.

To effectively leverage these three guidance signals described above for diffusion-based reconstruction, we propose DPCD, which integrates a diffusion model with two guidance paths to constrain the denoising process. In the first path, the text caption and RSRTs are fused into a unified semantic condition and injected into the diffusion U-Net via cross-attention, enabling semantic constraints to guide spatial features during denoising. In the second path, a trainable ControlNet\cite{zhang2023adding} extracts multi-scale visual features from the HCI and injects them into the diffusion model through additive residuals. By assigning each information stream a conditioning pathway suited to its representation, this design enables the structured integration of the three guidance signals while reducing potential interference among them.

Through the synergistic collaboration of these components, SDGIC achieves superior performance, as shown in Fig.\,\ref{fig:fig1}(e). In summary, the main contributions of this work are as follows:
\begin{itemize}
    \setlength{\itemsep}{0pt}
    \setlength{\parsep}{1pt}
    \setlength{\parskip}{0pt}
    \setlength{\topsep}{2pt}

    \item We formulate ultra-low-bitrate diffusion-based image compression as a semantic disambiguation problem and propose SDGIC, which uses a text caption (global semantic anchor), an HCI (dense visual anchor), and RSRTs (residual semantic constraints) to jointly reduce semantic ambiguity.

    \item We introduce RSRTs to encode reconstruction-relevant cues that remain ambiguous after conditioning on the text caption and HCI. Unlike generic semantic tokens or proxy-optimized representations, RSRTs are optimized toward the downstream denoising objective, making them aware of source-consistent reconstruction.

    \item To effectively integrate these three guidance signals, we design DPCD equipped with a dual-path conditioning mechanism. By decoupling semantic injection (via cross-attention with text and RSRTs) from spatial guidance (via a ControlNet branch with HCI), DPCD precisely constrains the generation process.

    \item Extensive experiments on multiple public benchmarks demonstrate that SDGIC achieves competitive semantic consistency at ultra-low bitrates while maintaining favorable perceptual quality.
\end{itemize}

The remainder of this paper is organized as follows. Section~\ref{sec:related_work} reviews related work. Section~\ref{sec:methodology} details the proposed framework and its key components. Experimental results and analysis are presented in Sections~\ref{sec:experiments} and~\ref{sec:discussion}, respectively. Finally, Section~\ref{sec:conclusion} concludes the paper.

\section{Related Work}
\label{sec:related_work}
\subsection{From Traditional to VAE-based Image Compression}
Traditional codecs (e.g., JPEG\cite{wallace1991jpeg}, BPG\cite{bellard2015bpg}, and VCC\cite{bross2021overview}) are built upon handcrafted, block-based coding strategies. Despite their long-standing optimization, the inherent limitations of such local processing inevitably lead to visually unpleasant artifacts like blocking and ringing, which become particularly problematic at ultra-low bitrate.

Advances in deep learning, particularly deep neural networks\cite{he2022elic,wang2025task,cao2026efficient,liu2024efficient,mishra2022deep}, have enabled VAE-based image compression\cite{feng2025linear,jiang2025mlic++,li2023frequency,zeng2025mambaic} to surpass traditional codecs on multiple benchmarks. Fu et al.\cite{fu2023learnedtip} proposed a discretized Gaussian-Laplacian-Logistic mixture model, combined with a concatenated residual blocks structure, achieving superior performance. Bao et al.\cite{bao2025stable} proposed a coherent demodulation-based transformation to improve the successive re-compression stability of neural image compression. Li et al.\cite{li2025learned} proposed a novel Hierarchical Progressive Context Model. Lu et al.\cite{lu2025learned} proposed a dictionary-based cross-attention entropy model, employing a learnable dictionary to capture structural priors from training data. However, these methods, optimized for pixel fidelity, often disrupt the statistical distribution alignment, sacrificing visual realism through texture loss and over-smoothing at ultra-low bitrate.

\subsection{Generative Image Compression}
Generative image compression shifts the objective from pixel fidelity to semantic consistency. It achieves high-quality reconstruction through extremely limited semantic guidance. For GANs-based methods, Mentzer et al.\cite{mentzer2020high} proposed HiFiC that inputs quantized latent representations into both the generator and discriminator for optimization. Muckley et al.\cite{muckley2023improving} proposed a non-binary discriminator based on quantized local image representations obtained from a VQ-VAE autoencoder. Such GAN-based methods often suffer from training instability and mode collapse, and are prone to generating structural artifacts that deviate from the original semantic content.

Diffusion models have demonstrated significant potential in image compression, offering a new technical pathway for generative compression\cite{xue2026one}. Lei et al.\cite{lei2023text+} achieved high-quality reconstruction using only text prompts and sketches via a pre-trained model. Li et al.\cite{li2026structure} proposed FPD-IC, a two-stage conditional diffusion framework for balancing structural fidelity and perceptual quality.
Gao et al. proposed M-CMC\cite{gao2024rate}, which decouples images into structure and texture, utilizing a novel reinforcement learning approach to optimize rate-distortion performance. 
In another study, Li et al.\cite{li2024misc} presented a framework called MISC that integrated LMM into the codec for the first time.
Gao et al. introduced MKIC\cite{gao2025exploring}, which provides valuable insights by combining natural visual knowledge with human language to achieve high-quality reconstruction.
Careil et al.\cite{careil2023towards} combined vector-quantized latent representations and text to condition the generation process. However, at ultra-low bitrates, these methods often provide incomplete guidance for diffusion-based reconstruction, leaving the source content underconstrained and causing semantic deviation.

Other attempts have focused on compressing images into single-modal latent representations to guide diffusion models. Yang et al.\cite{yang2023lossy} mapped images into contextual latent variables to guide reconstruction. Li et al.\cite{li2024towards} combined VAEs with pre-trained diffusion models, and leveraged latent features to enhance fidelity. 
However, such methods may over-rely on single low-level CNN features, which can restrict the generative capacity of the diffusion model and consequently compromise perceptual quality.

\subsection{Diffusion Models as Generative Priors}
The core idea of diffusion models is to generate high-quality samples through a learned, progressive denoising process. Although proposed earlier\cite{sohl2015deep}, diffusion models only gained broad recognition\cite{kingma2021variational, ren2025your,yang2025conditional} after Ho et al.\cite{ho2020denoising} introduced the Denoising Diffusion Probabilistic Model (DDPM), which simplified the training objective and exhibited powerful generative performance. Subsequently, Rombach et al.\cite{rombach2022high} proposed the Latent Diffusion Model (LDM), moving the diffusion process from pixel space to a compact latent space. This shift greatly reduced computational cost and paved the way for large-scale real-world applications such as Stable Diffusion.

The framework of diffusion models comprises a fixed forward noising process and a learnable reverse denoising process. In the forward phase, data $\mathbf{x}_0$ is progressively perturbed by Gaussian noise via a Markov chain, given by:
\begin{equation}
q(x_t \mid x_{t-1}) = \mathcal{N}\!\big(x_t;\, \sqrt{1-\beta_t}\,x_{t-1},\, \beta_t \mathbf{I}\big).
\label{eq:forward}
\end{equation}
with $\beta_t$ as the noise schedule. The core learning task is the reverse process, where a network is trained to predict the noise present at any timestep $t$, allowing the model to generate high-quality samples from pure noise via iterative denoising.

Diffusion models are uniquely suited for ultra-low bitrate compression due to their powerful generative priors derived from large-scale training, which compensate for severe information loss at the decoder under ultra-low bitrate constraints. They can synthesize perceptually important details from sparse guidance, enhancing reconstruction realism. This ``strong prior, weak guidance" paradigm aligns well with the goal of high perception ultra-low bitrate compression. However, this approach introduces a central challenge: ensuring that the prior-driven generation remains semantically faithful to the original content. Addressing this semantic consistency issue is the core motivation behind our proposed SDGIC framework.

\section{Methodology}
\label{sec:methodology}
\subsection{Architecture Overview and Information Flow}

\begin{figure*}[t] 
    \centering
    \includegraphics[width=\textwidth]{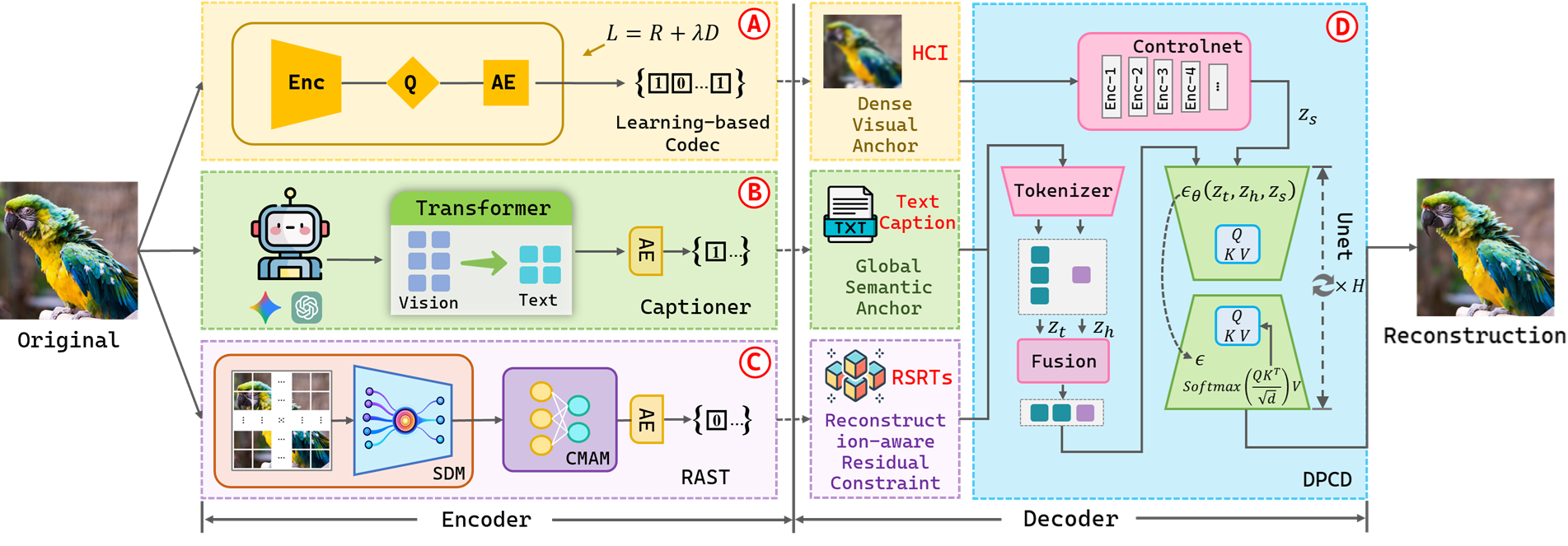} 
    \caption{Overall framework of the proposed SDGIC. The encoder transforms the image into three complementary guidance streams, including a text caption, an HCI, and RSRTs generated by the RAST. The text caption and RSRTs are losslessly compressed into bitstreams using Zstd\cite{collet2021rfc}, while the HCI is compressed via arithmetic coding. The decoder employs a DPCD, which integrates these guidance streams with the generative prior of the diffusion model to reconstruct the image.}
    \label{fig:fig2}
\end{figure*}

Fig.\,\ref{fig:fig2} shows the overall architecture of the proposed SDGIC framework. At the encoder side, the original image is transformed into three guidance streams: a concise caption generated by a text captioner, an HCI produced by a learned image codec, and RSRTs generated by the RAST. RAST consists of a Semantic Refinement Module (SRM) and a Cross-Modal Alignment Module (CMAM), which distill reconstruction-relevant semantics from continuous visual representations and align them into a text-compatible embedding space, respectively. At the decoder side, the caption and RSRTs are fused and injected into the diffusion model through cross-attention, whereas the HCI is injected through a trainable ControlNet branch in the form of multi-scale residuals. By combining these three guidance streams with the generative prior of the diffusion model, DPCD reconstructs high-quality images.

The methodologies and roles of these three guidance streams are detailed in Section\,\ref{sec:3.2}. The design specifics of the DPCD are elaborated in Section\,\ref{sec:3.3}, while the training strategy and loss functions are introduced in Section\,\ref{sec:3.4}.

\subsection{Semantic Disambiguation Guidance} \label{sec:3.2}
\vspace{2pt}
\subsubsection{Text Caption for Global Semantic Anchoring}
\vspace{2pt}
Despite being a cost-effective guidance signal in generative compression, text is limited by the intrinsic modality gap between discrete tokens and continuous image scenes. This gap hinders perfect text-image alignment, as the coarse granularity of text fails to accurately map fine-grained visual details. To formalize this issue, we encode the image $x$ and the text caption $z_{\text{text}}$ into embeddings $v_I$ and $v_T$ using encoders, $E_I$ and $E_T$ , as follows:
\begin{equation}
\left\{
\begin{aligned}
v_I &= E_I(x), \\
v_T &= E_T(z_{\text{text}}).
\end{aligned}
\right.
\label{eq:vI_reformatted}
\end{equation}

Due to its discrete and symbolic nature, text provides only an abstract, lossy summary of the rich and continuous visual scene. Consequently, the embeddings $v_I$ and $v_T$, projected from these intrinsically mismatched sources, cannot achieve perfect alignment in the shared latent space, which results in an invariably positive cosine distance, as follows:
\begin{equation}
d_{\cos}(v_I, v_T) = 1 - \frac{|v_I \cdot v_T|}{\|v_I\| \cdot \|v_T\|} > 0.
\label{eq:cosine}
\end{equation}

Note that over-detailed descriptions cannot close the modality gap and may introduce semantic noise. Thus, in SDGIC, the role of text caption is redefined as a concise global semantic anchor rather than a carrier of fine-grained visual details. A captioner (e.g., an online LMM, BLIP-2\cite{li2023blip}, or lightweight GIT\cite{wang2022git}) generates a caption under the $P_{\text{prompt}}$: \textit{``Extract and concisely articulate the core, unambiguous semantic information of this image (main subject and context), in under 20 words.''} The caption is then losslessly compressed using Zstd. This design prevents semantic drift at the macro level in reconstruction. The influence of text caption and its length will be quantitatively analyzed in Section\, \ref{sec:discussion}.

\subsubsection{HCI for Dense Visual Anchoring}
\vspace{2pt}
Existing methods often use sparse representations, such as sketches or semantic maps, to guide image reconstruction. However, these representations typically preserve only selective structural features, e.g., contours, resulting in low information density and incomplete visual evidence. Although such guidance helps generative models synthesize visually rich details and improve perceptual quality, its limited coverage leaves many aspects of the source image underconstrained, thereby increasing semantic ambiguity. Therefore, constructing guidance representations that retain denser and more stable source-specific information is crucial for reducing semantic ambiguity and better anchoring the generation process.

\begin{figure}[pos=t]
    \centering
    \includegraphics[width=0.92\columnwidth]{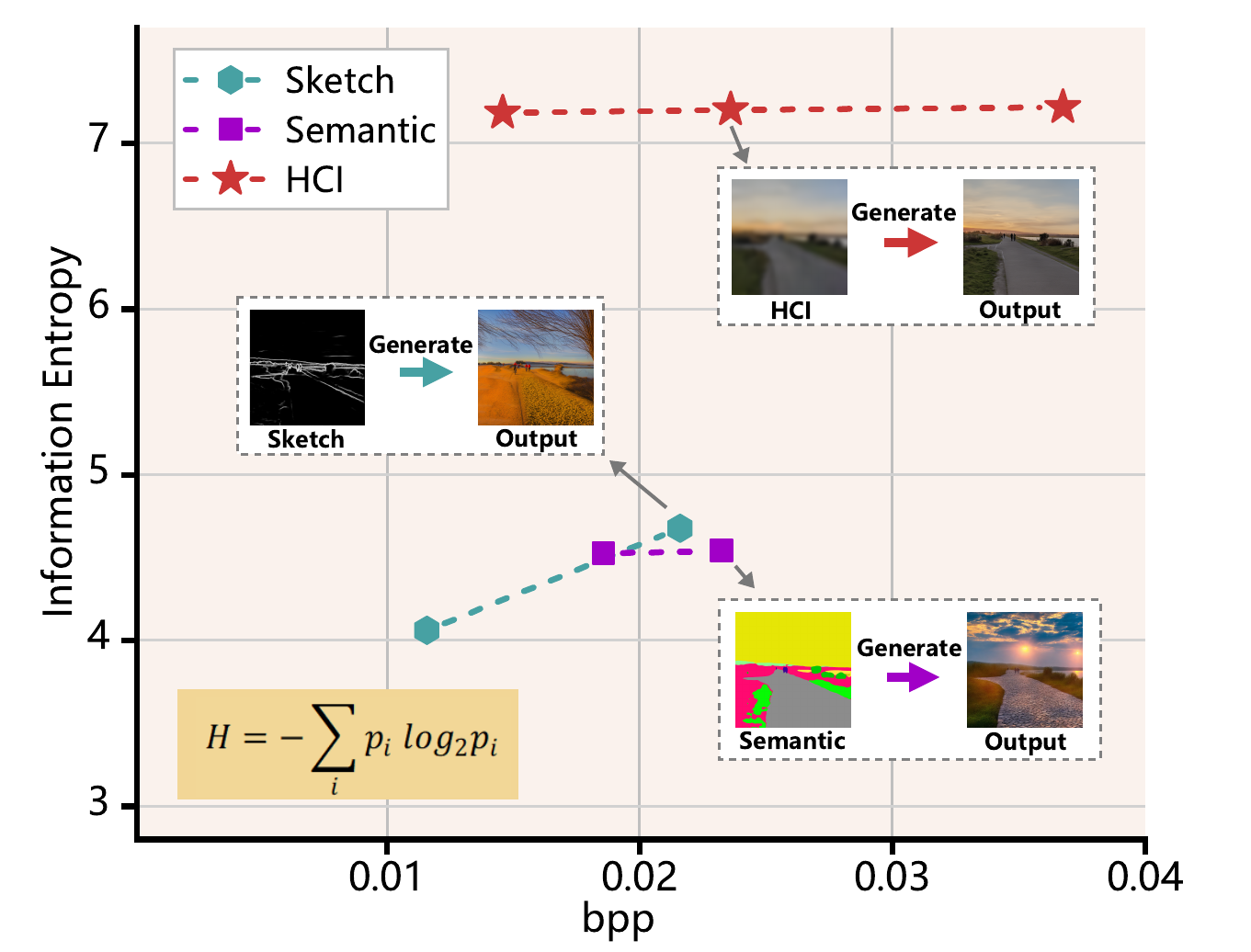}
    \caption{Information entropy comparison: HCI vs. sketch and semantic map (denoted as ``Semantic'' in the figure). HCI exhibits significantly higher information entropy, demonstrating its superior information density.}
    \label{fig:fig3}
\end{figure}

In the proposed SDGIC framework, we introduce a learning-based image codec, ELIC\cite{he2022elic}, owing to its excellent trade-off between compression performance and computational latency. It generates an HCI that serves as a dense visual anchor for diffusion-based reconstruction. Unlike existing sparse guidance strategies, HCI does not rely on manually designed features. By adhering to the rate--distortion principle:
\begin{equation}
L = R + \lambda \cdot D,
\label{eq:rate-distortion}
\end{equation}
the codec preserves visual information through efficient global bit allocation under limited bitrates. Here, $R$ is the bitrate, $D$ is the distortion, and $\lambda$ is the Lagrange multiplier.
At equivalent bitrates, HCI exhibits higher information density and retains richer low-level visual evidence than sparse representations. This advantage enables it to provide stronger constraints during image generation, thereby effectively suppressing structural distortion and color artifacts.

To quantitatively assess information density, we use image information\cite{jeon2021information}:
\begin{equation}
H = -\sum_i p_i \log_2 p_i,
\end{equation}
where \( p_i \) denotes the probability of pixel value \( i \). A higher \( H \) indicates greater information density. As shown in Fig.\,\ref{fig:fig3}, HCI achieves 57.17\% and 58.88\% higher entropy than two commonly used sparse representations at the same bitrate.

The process of generating the HCI $\hat{x}$ is formulated as:
\begin{equation}
\left\{
\begin{aligned}
&B(x)= S \left( Q \left( E_c(x) \right) \right), \\
&\hat{x} = D_c \left( S^{-1} \left( B(x) \right) \right).
\end{aligned}
\right.
\label{eq:codec_process}
\end{equation}
here, $E_c$ is the encoder that maps the input image $x$ into a latent representation, which is subsequently quantized by $Q$ and entropy-coded by $S$ into the bitstream $B(x)$. The decoding process involves entropy decoding $S^{-1}$ and a decoder $D_c$, reconstructing the HCI $\hat{x}$ from the bitstream.

This HCI can be regarded as an informational substrate that provides dense low-level visual priors, such as spatial layout, structural, color distributions, and tonal cues. These priors strongly anchor the generative process, guiding the model to synthesize details in a constrained and source-faithful manner. Section\,\ref{sec:discussion} will quantitatively discuss the effectiveness of this strategy based on performance metrics.

\subsubsection{RSRTs for Reconstruction-Relevant Semantic Constraints}
\vspace{2pt}
As discussed above, the central challenge in generative image compression at ultra-low bitrates is the semantic ambiguity induced by insufficient conditioning. Text captions provide global semantic cues, while HCI preserves dense low-level visual evidence; however, there still exist source-specific reconstruction cues that are not fully captured by either stream. If these residual cues are absent, multiple plausible reconstructions may satisfy the same caption and HCI conditions, allowing the diffusion prior to generate visually reasonable but source-inconsistent details.

\begin{figure}[pos=t]
    \centering
    \includegraphics[width=0.9\linewidth]{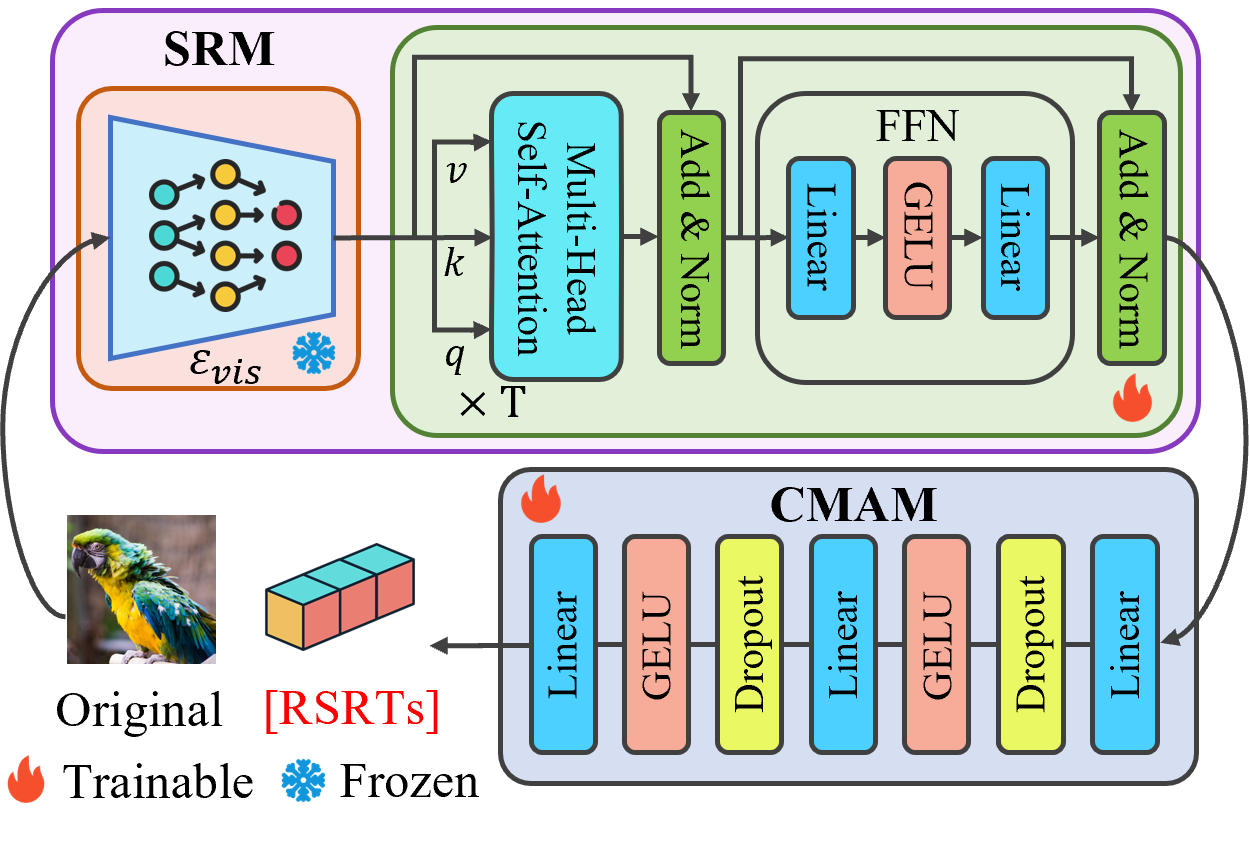} 
    \caption{\!Detailed architecture of the RAST, comprising an SRM and a CMAM, and optimized through the downstream denoising objective for reconstruction-aware semantic token generation.}
    \label{fig:fig4}
\end{figure}

To address this, we introduce RSRTs, which serve as compact source-specific semantic constraints complementary to the text caption and HCI. The RSRTs are generated by RAST, as shown in Fig.\,\ref{fig:fig4}. Rather than encoding generic visual semantics, the tokenizer is optimized through the downstream image reconstruction objective, enabling the learned tokens to focus on reconstruction-relevant information that helps distinguish the source-consistent image from other plausible candidates.

Given an input image \(x \in \mathbb{R}^{H \times W \times 3}\), a frozen pre-trained visual encoder \(\epsilon_{\text{vis}}\) first maps the image into a continuous visual representation:
\begin{equation}
    Z^{(0)} = \epsilon_{\text{vis}}(x) \in \mathbb{R}^{N \times D},
\end{equation}
where \(N\) denotes the number of visual tokens and \(D\) denotes the embedding dimension. The visual encoder is instantiated with OpenCLIP ViT-H/14, whose cross-modal pretraining provides rich semantic representations. Since the encoder itself is frozen, its output is further processed by a trainable semantic refinement module to adapt the visual representation to the downstream reconstruction objective.

Specifically, the semantic refinement module refines the full visual token sequence \(Z^{(0)}\) through \(T=2\) trainable transformer-style self-attention layers, each with 16 attention heads. Compared with local convolutional operations, such a self-attention design provides stronger capability for capturing long-range semantic interactions. At the \(t\)-th layer, the self-attention operation is formulated as:
\begin{equation}
    Y^{(t)} = \text{Softmax}\left( \frac{Q^{(t)}(K^{(t)})^T}{\sqrt{d_k}} \right) V^{(t)} ,
\end{equation}
where \(Q^{(t)}\), \(K^{(t)}\), and \(V^{(t)}\) are the query, key, and value projections at the \(t\)-th layer, and \(d_k\) is the scaling factor for numerical stability. Each layer is followed by residual connection, layer normalization, and a feed-forward network, producing the layer output \(Z^{(t)}\). After \(T\) layers, the refined visual token sequence \(Z^{(T)}\) is obtained. The final CLS token \(z_{\mathrm{CLS}}^{(T)}\) is then extracted from \(Z^{(T)}\) as a compact reconstruction-aware semantic representation for subsequent cross-modal alignment. All trainable parameters within this \(T\)-layer self-attention network are collectively denoted as \(\phi\).

The refined representation is then compressed and projected into the textual conditioning space by a lightweight cross-modal alignment module parameterized by \(\varphi\):
\begin{equation}
    Z_{\text{RSRT}} = \text{MLP}_{\varphi}\left(z_{\mathrm{CLS}}^{(T)} \right).
    \label{eq:RSRT_simplified}
\end{equation}
where \(Z_{\text{RSRT}} \in \mathbb{R}^{D_{\text{text}} \times L}\), \(L\) denotes the number of RSRTs, and \(D_{\text{text}}\) is the textual embedding dimension. By aligning RSRTs with the textual embedding space, they can be seamlessly fused with the caption condition and injected into the diffusion decoder through cross-attention. 

Through this reconstruction-aware optimization, RSRTs learn to preserve source-specific semantic cues that are not sufficiently conveyed by either the caption or HCI. More importantly, the term “residual” here refers to functional complementarity rather than explicit feature subtraction. In this way, RSRTs provide additional constraints for source-consistent reconstruction. Their efficacy will be qualitatively and quantitatively validated in Section\, \ref{sec:discussion}.

\subsection{Dual-Path Conditioned Diffusion Decoder} \label{sec:3.3}
\begin{figure}[pos=t]
    \centering
    \includegraphics[width=0.485\textwidth]{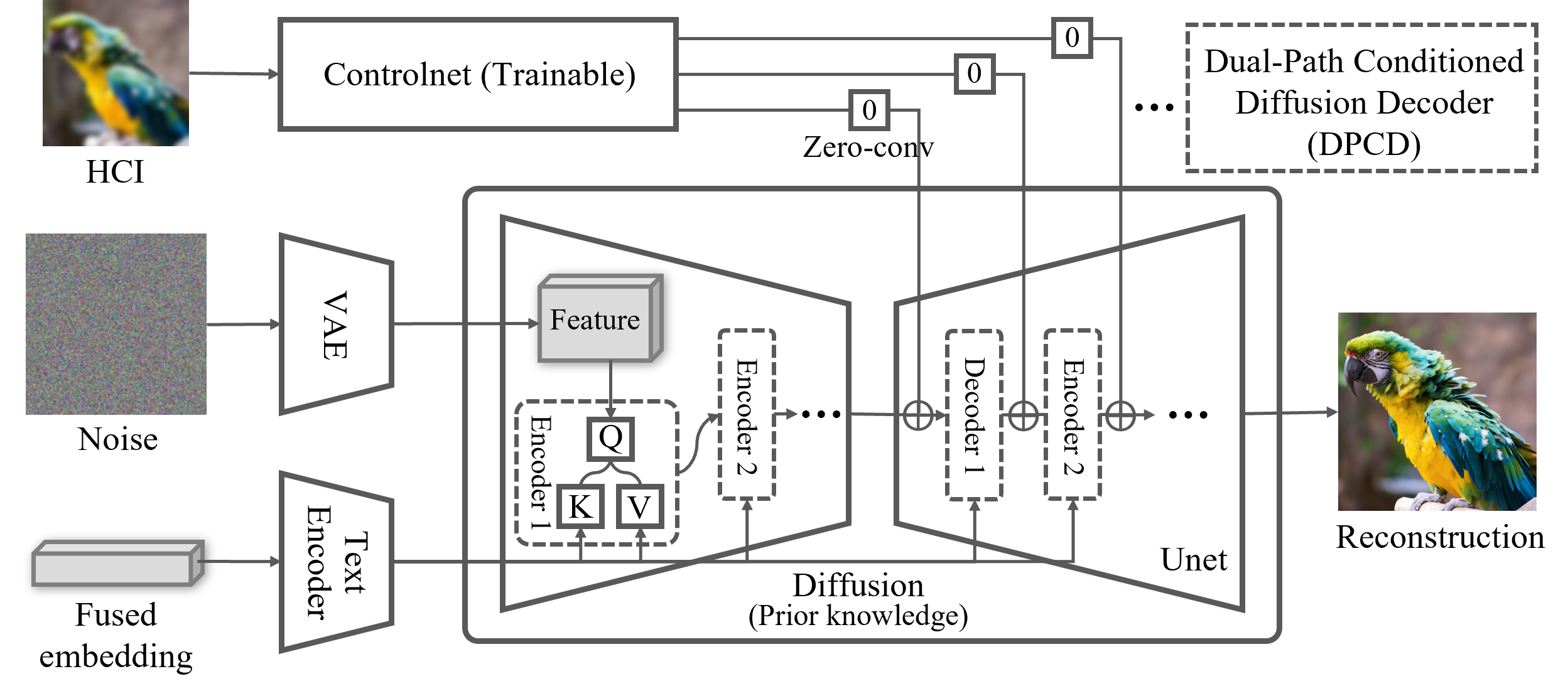}
    \caption{\!Detailed architecture of DPCD, leveraging a pre-trained diffusion model with dual-path conditioning that combines cross-attention and ControlNet additive residuals for reconstruction guidance.}
    \label{fig:fig5}
\end{figure}

\vspace{2pt}
To effectively leverage the three guidance streams for image reconstruction, we design DPCD, as shown in Fig.\,\ref{fig:fig5}. Built upon a pre-trained diffusion model, DPCD exploits the generative prior of the diffusion backbone while using the text caption, HCI, and RSRTs to reduce semantic ambiguity during iterative denoising. Since these guidance streams differ in representation form and conditioning role, directly fusing them into a single conditioning channel may cause modal interference and weaken their respective constraints. Therefore, DPCD adopts a dual-path conditioning mechanism that assigns them to injection pathways suited to their characteristics and roles.

In the first path, the text caption and RSRTs are fused into a unified semantic condition and injected into the denoising U-Net through cross-attention. Specifically, inspired by Gal et al.\cite{gal2022image}, we insert predefined placeholder tokens into the text caption and then replace the corresponding token embeddings with the RSRT embeddings generated by RAST. The resulting fused embedding sequence is processed by the diffusion model's text encoder to form a unified semantic condition. During generation, this condition acts as the Key and Value matrices, interacting with the U-Net's intermediate feature maps as the Query through cross-attention layers. This mechanism enables semantic concepts to be mapped to specific spatial locations in the feature maps, thereby providing precise semantic guidance for the denoising process.

In the second path, inspired by Zhang et al.\cite{zhang2023adding}, we introduce a parallel ControlNet module to steer the denoising process using the HCI as a spatial condition. Specifically, the trainable ControlNet extracts multi-scale features from the HCI to capture rich low-level visual information, including structure, layout, color, and appearance cues. These features are subsequently added as residuals into the corresponding middle and decoder blocks of the frozen main U-Net. This parallel architecture with residual injection ensures that the main U-Net retains its generative prior intact while receiving explicit spatial guidance from the compressed image at each denoising step.

Through this dual-path conditioning mechanism, DPCD precisely injects semantic and spatial guidance, reducing generation ambiguity at ultra-low bitrates and improving reconstruction quality.

\subsection{Training Strategy and Loss Function} \label{sec:3.4}
\vspace{2pt}
This subsection describes the training strategy for the proposed framework. Directly training the entire pre-trained diffusion model typically suffers from slow convergence and high computational cost. To address this, we freeze the diffusion backbone and optimize only the trainable guidance modules, including RAST and the ControlNet branch. In this way, SDGIC adapts the semantic and spatial guidance modules to the downstream reconstruction objective while avoiding full-model training. This design follows the spirit of parameter-efficient adaptation, but our focus is to optimize the guidance modules.

Furthermore, we employ a three-stage strategy to train the model. Specifically,

\noindent\textbf{Stage I.}
We disable RAST and train only the ControlNet module to ensure robust spatial conditioning. The optimization objective is defined as:
\begin{equation}
L_{\text{I}}(\psi) = \mathbb{E}_{\substack{t, x_0, z_t, z_h, \epsilon \sim \mathcal{N}(0, \mathbf{I})}}\! \left[ \left\| \epsilon\! - \!\epsilon_\theta(x_t, \!t,\! z_t, \!z_h; \psi) \right\|_2^2 \right],
\label{eq:stage1}
\end{equation}
where $z_t$ and $z_h$ denote the text caption and the HCI conditioning, respectively. $\psi$ represents the trainable parameters of ControlNet, $x_t$ is the noisy latent variable at timestep $t$.

\noindent\textbf{Stage II.}
We freeze the ControlNet weights, introduce RAST to be trained, allowing it to learn reconstruction-aware residual semantic constraints independently. The optimization objective is defined as:
\begin{equation}
\begin{aligned}
& L_{\text{II}}(\varphi, \phi) = \\ 
& \mathbb{E}_{\substack{t, x_0, z_t, z_h, z_r, \epsilon \sim \mathcal{N}(0, \mathbf{I})}} 
\left[ 
\left\| 
\epsilon \!- \!\epsilon_\theta(x_t, t, z_t, z_h, z_r; \varphi, \phi) 
\right\|_2^2 
\right],
\end{aligned}
\label{eq:stage2}
\end{equation}
where $z_r$ represents the RSRT conditioning. $\varphi$ and $\phi$ denote the trainable parameters of the SRM and the CMAM within RAST, respectively.

\noindent\textbf{Stage III.}
We jointly fine-tune both the RAST and the ControlNet parameters to achieve optimal synergistic interplay between the multimodal guidance signals. The optimization objective is defined as:
\begin{equation}
\begin{aligned}
& L_{\text{III}}(\varphi, \phi, \psi) = \\ 
& \mathbb{E}_{t, x_0, z_t, z_h, z_r, \epsilon \sim \mathcal{N}(0, \mathbf{I})} 
\!\!\left[
\left\| 
\epsilon \!- \!\epsilon_\theta(x_t, \!t,\! z_t,\! z_h,\! z_r; \!\varphi, \!\phi, \!\psi) 
\right\|_2^2
\right].
\end{aligned}
\label{eq:stage3}
\end{equation}

This staged training strategy enhances both training efficiency and stability without compromising the generative capability of the model

\section{Experiments}
\label{sec:experiments}
\subsection{Experiment Settings}
\vspace{2pt}
\noindent\textbf{Datasets.}
During the training phase, we construct the training set using 166,451 image-text pairs from the CC3M\cite{sharma2018conceptual} dataset, with all images randomly cropped to 512$\times$512 pixels. To evaluate our model, we employ three widely-used benchmark datasets at different scales: Tecnick, CLIC2020, and DIV2K. Specifically, all 140 images in the Tecnick dataset are uniformly rescaled and center-cropped to a fixed resolution of 768$\times$768 pixels. The 178 images from the CLIC2020 dataset are processed similarly, resulting in a uniform resolution of 1024$\times$1024 pixels. Additionally, 100 images from the DIV2K dataset are center-cropped to 1352$\times$1352 pixels for testing.

\vspace{2pt}
\noindent\textbf{Metrics.}
To comprehensively evaluate the quality of reconstructed images, we employ metrics covering semantic consistency, perceptual quality, and pixel-level fidelity. For semantic consistency, we use LPIPS \cite{zhang2018unreasonable}, DISTS\cite{ding2020image}, ClipScore and SeSS\cite{fan2024semantic}. For perceptual quality, we use CLIPIQA\cite{wang2023exploring} and MANIQA\cite{yang2022maniqa}. For pixel-level fidelity, we report PSNR and MS-SSIM\cite{wang2003multiscale}. Additionally, we adopt AFINE\cite{chen2025toward}, a state-of-the-art comprehensive image quality assessment metric.

\vspace{2pt}
\noindent\textbf{Baselines.}
To establish performance baselines, we select representative image compression methods spanning multiple categories. For traditional handcrafted codecs, we include the widely-used JPEG\cite{wallace1991jpeg} and VVC\cite{bross2021overview}. The VVC anchor is implemented using the official VTM software (version 23.13). For learning-based neural codecs, we incorporate ELIC\cite{he2022elic}, DCAE\cite{lu2025learned}, FTIC\cite{li2023frequency}, LALIC\cite{feng2025linear} and MLIC++\cite{jiang2025mlic++}. For GAN-based generative compression, we incorporate HiFiC\cite{mentzer2020high} and VQIR\cite{wei2024toward}. In the domain of diffusion-based generative compression, we consider: PICS\cite{lei2023text+}, PerCo\cite{careil2023towards}, MISC\cite{li2024misc} and DDCM\cite{ohayon2025compressed}. 

\vspace{2pt}
\noindent\textbf{Specific Implementation.}
In our implementation, we employ Gemini 2.5 Pro to extract concise text captions, and build the DPCD on Stable Diffusion 2.1-base. For the ultra-low bitrate setting (bpp\,$<$\,0.05), we use a caption of approximately 20 words, corresponding to about 0.002 bpp, and set the number of RSRTs to 1 with a dimension of 1$\times$1024. To evaluate performance at different target bitrates, we train three model variants by setting the Lagrange multiplier $\lambda_{\mathrm{ELIC}}$ in the ELIC compressor to $\{1,2,3\}\times10^{-4}$, respectively. As shown in Table\,\ref{tab:bpp_comparison}, the bitrate allocation of each guidance stream is reported on the CLIC2020 dataset.
The model is optimized using the Adam optimizer with $\beta_1=0.9$ and $\beta_2=0.999$. The learning rate is set to $1\times10^{-4}$ in Stages I and II, and reduced to $5\times10^{-5}$ in Stage III. All experiments are conducted on an NVIDIA A40 GPU with a batch size of 16. Each training stage runs for 8 epochs, totaling approximately 83k iterations. We use full precision for training and half precision for inference to accelerate reconstruction and reduce memory overhead. During inference, images are reconstructed with 50 denoising steps.
\begin{table}[t]
    \vspace{2pt}
    \centering
    \caption{Average bpp of different guidance streams on CLIC2020.}
    \label{tab:bpp_comparison}
    \renewcommand{\arraystretch}{1.24}
    \setlength{\arrayrulewidth}{0.5pt}
    \setlength{\tabcolsep}{2.4pt}
    {\rmfamily\fontsize{9pt}{10.8pt}\selectfont
    \begin{tabular}{c|ccccc}
        \noalign{\hrule height 1.0pt}
        \multirow{2}{*}{\textbf{Guidance}}
        & \multirow{2}{*}{\shortstack{Text caption}}
        & \multirow{2}{*}{RSRTs}
        & \multicolumn{3}{c}{HCI ($\lambda_{\mathrm{ELIC}}$)} \\
        \cline{4-6}
        & & & $1{\times}10^{-4}$ & $2{\times}10^{-4}$ & $3{\times}10^{-4}$ \\
        \hline
        \rowcolor{lightblue}
        \textbf{bpp} & 0.002 & 0.014 & 0.014 & 0.023 & 0.036 \\
        \noalign{\hrule height 1.0pt}
    \end{tabular}
    }
    \vspace{2pt}
\end{table}

\subsection{Experiment Results}
\vspace{2pt}
\begin{figure*}[pos=t] 
    \centering
    \includegraphics[width=0.98\textwidth]{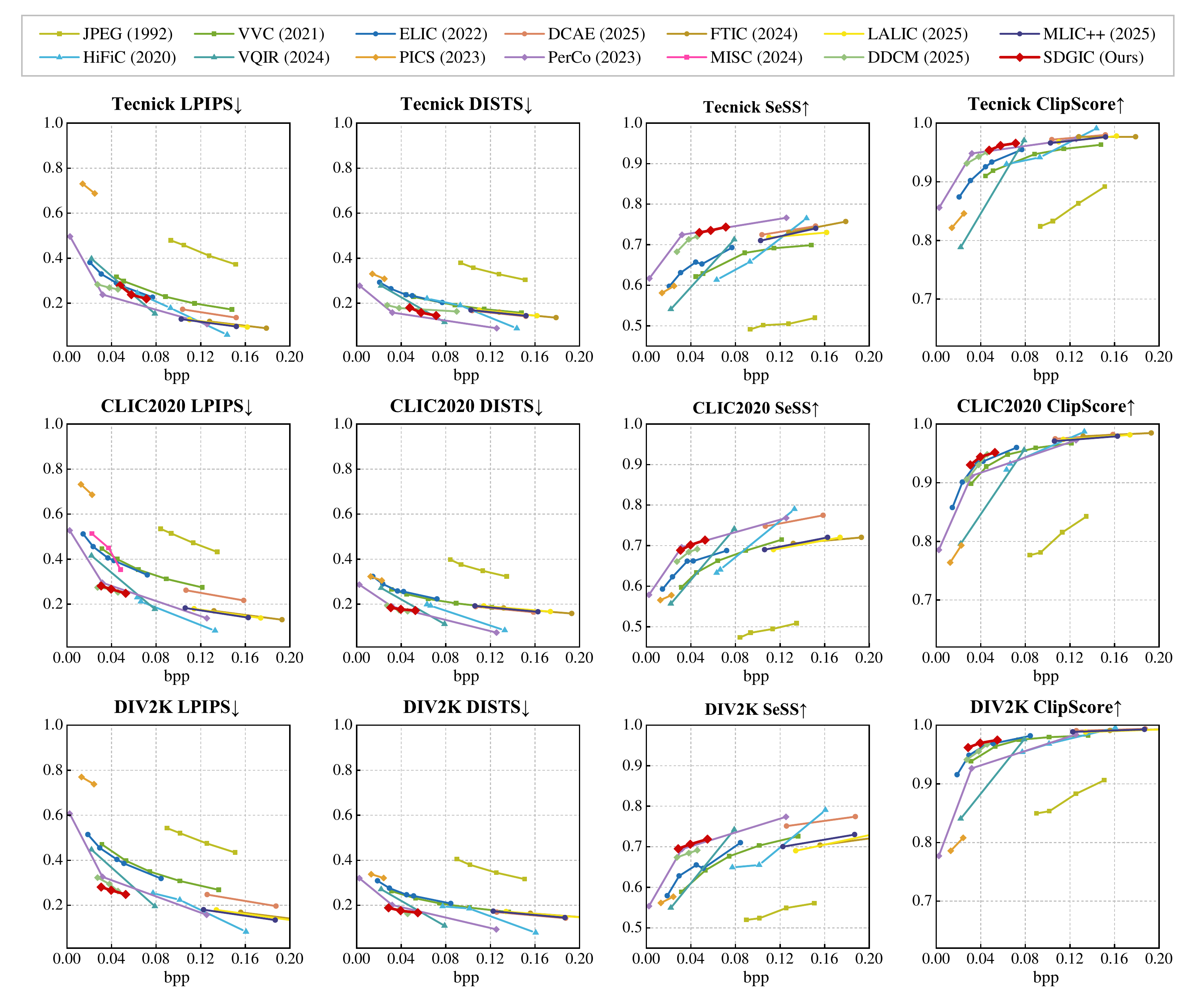} 
    \caption{Quantitative comparison of semantic consistency metrics (LPIPS, DISTS, SeSS and CLIPScore) evaluated on the Tecnick, CLIC2020 and DIV2K datasets at ultra-low bitrate. Our method achieves highly competitive performance compared with both conventional and generative baselines (JPEG\cite{wallace1991jpeg}, VVC\cite{bross2021overview}, ELIC\cite{he2022elic}, DCAE\cite{lu2025learned}, FTIC\cite{li2023frequency},  LALIC\cite{feng2025linear}, MLIC++\cite{jiang2025mlic++}, HiFiC\cite{mentzer2020high}, VQIR\cite{wei2024toward}, PICS\cite{lei2023text+}, PerCo\cite{careil2023towards}, MISC\cite{li2024misc}), DDCM\cite{ohayon2025compressed}.}
    \label{fig:fig6}
\end{figure*}
\begin{figure*}[pos=t] 
    \centering
    \includegraphics[width=0.9\linewidth]{Fig/Fig7.png}
    \caption{Quantitative comparison of the comprehensive metric AFINE evaluated on CLIC2020, DIV2K, and Tecnick datasets at ultra-low bitrate. Our method demonstrates superior performance over other generative baselines (HiFiC\cite{mentzer2020high}, VQIR\cite{wei2024toward}, PICS\cite{lei2023text+}, PerCo\cite{careil2023towards}, DDCM\cite{ohayon2025compressed}).}
    \label{fig:fig7}
\end{figure*}
\begin{figure*}[t]
    \centering
    \includegraphics[width=\textwidth]{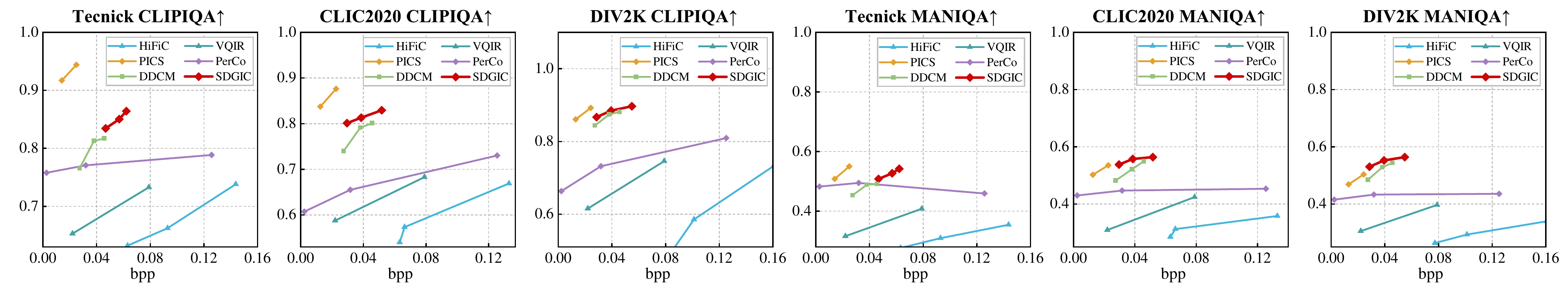} 
    \caption{Quantitative comparison of perceptual metrics (MANIQA, CLIPIQA) evaluated on CLIC2020, DIV2K, and Tecnick datasets with generative baselines (HiFiC\cite{mentzer2020high}, VQIR\cite{wei2024toward}, PICS\cite{lei2023text+}, PerCo\cite{careil2023towards}, DDCM\cite{ohayon2025compressed}). Our method demonstrates highly competitive performance in perceptual quality evaluation.}
    \label{fig:fig8}
\end{figure*}
\begin{figure*}[!ht]
    \centering
    \includegraphics[width=\textwidth]{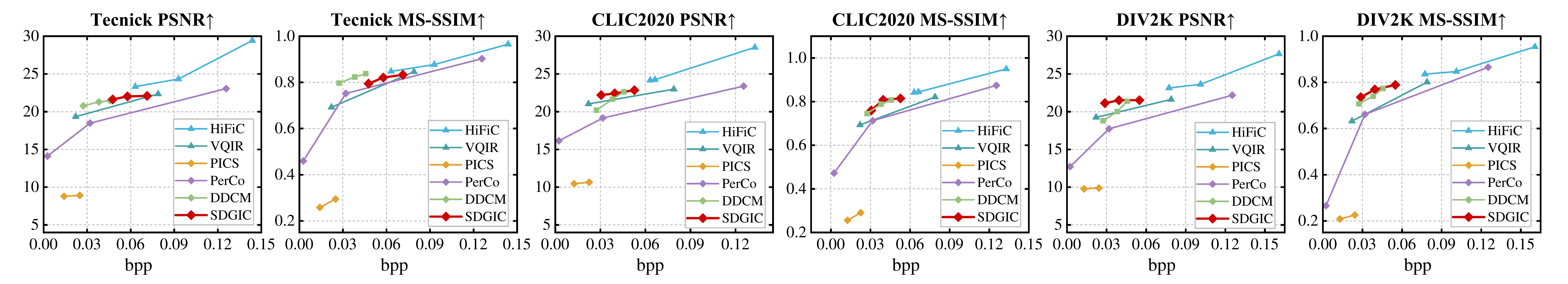} 
    \caption{Quantitative comparison of pixel-level fidelity metrics (PSNR, MS-SSIM) evaluated on CLIC2020, DIV2K, and Tecnick datasets with generative baselines (HiFiC\cite{mentzer2020high}, VQIR\cite{wei2024toward}, PICS\cite{lei2023text+}, PerCo\cite{careil2023towards}, DDCM\cite{ohayon2025compressed}). Our method also achieves highly competitive performance  in terms of pixel-level consistency.}
    \label{fig:fig9}
\end{figure*}
\begin{figure*}[t]
    \centering
    \includegraphics[width=0.96\textwidth]{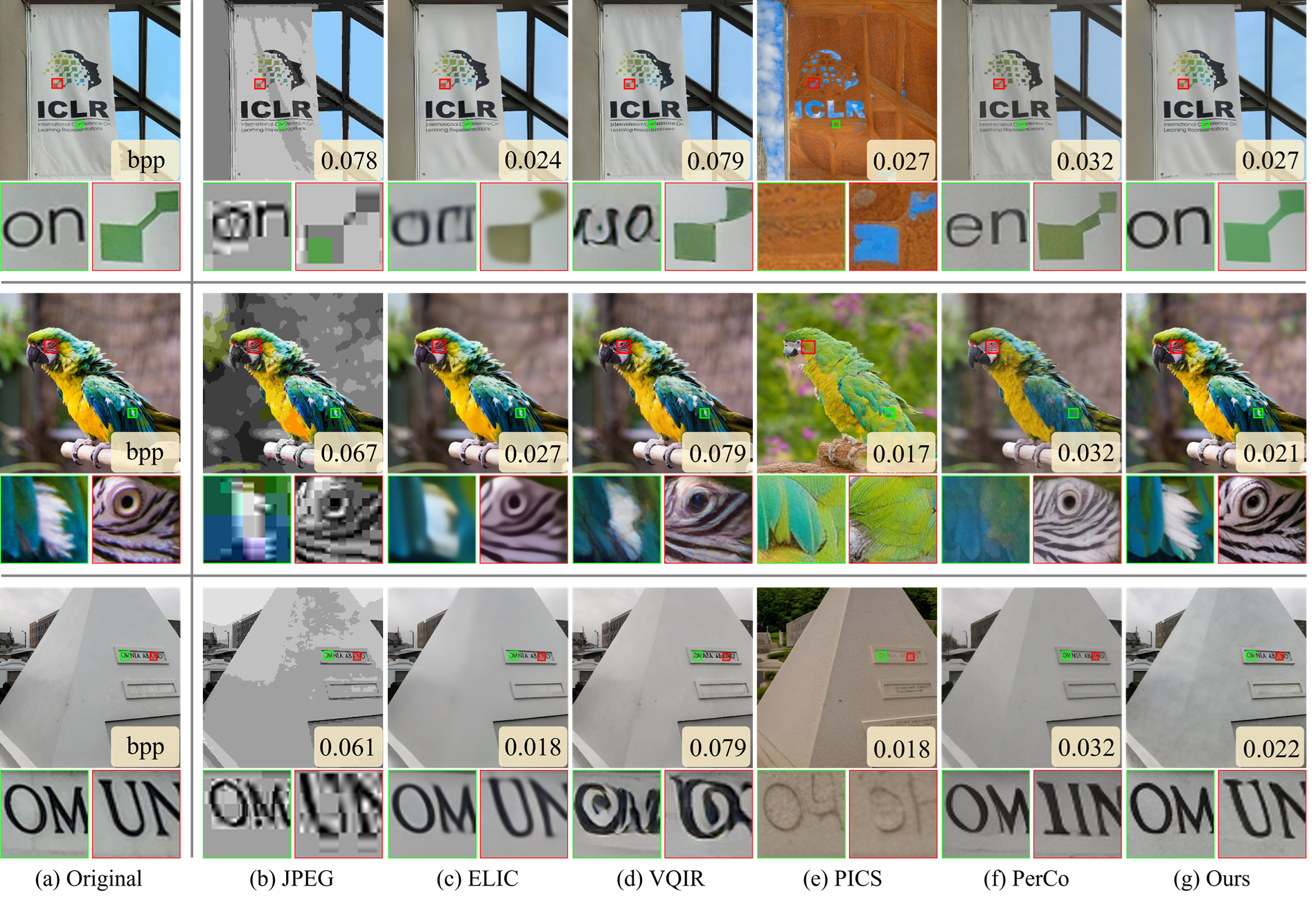} 
    \caption{Qualitative comparison with representative baselines (JPEG\cite{wallace1991jpeg}, ELIC\cite{he2022elic}, VQIR\cite{wei2024toward}, PICS\cite{lei2023text+}, PerCo\cite{careil2023towards}) on sample images from the test datasets at ultra-low bitrate, our method reconstructs images that are more realistic and more consistent with the original images.}
    \label{fig:fig10}
\end{figure*}

We conduct extensive experiments to evaluate the effectiveness of the proposed SDGIC. The results are organized as follows. Fig.\,\ref{fig:fig6} presents the comparison of semantic consistency metrics; Fig.\,\ref{fig:fig7} reports the comparison on the state-of-the-art comprehensive metric AFINE; Fig.\,\ref{fig:fig8} shows perceptual quality comparisons against representative generative baselines; and Fig.\,\ref{fig:fig9} reports the comparison on conventional pixel-level fidelity metrics. In addition, Fig.\,\ref{fig:fig10} provides qualitative comparisons of reconstructed images. It should be noted that conventional codecs and VAE-based codecs are optimized for pixel-level fidelity and usually produce over-smoothed reconstructions at ultra-low bitrates. Therefore, their results on MANIQA, CLIPIQA, and AFINE are not reported. For MISC, we compare with the results reported in its original paper, since its source code is unavailable.

As shown in Fig.\,\ref{fig:fig6}, SDGIC achieves the best overall performance in semantic consistency across the three public benchmark datasets. Conventional codecs, such as JPEG and VVC, suffer from severe blocking artifacts, blurring, and structural degradation at ultra-low bitrates, which weakens their semantic consistency. Similarly, VAE-based learned codecs, including ELIC, DCAE, FTIC, LALIC, and MLIC++, mainly rely on rate-distortion optimization. Under stringent bit budgets, these methods tend to discard high-frequency details and produce over-smoothed reconstructions due to the lack of generative priors, resulting in degraded performance (Fig.\,\ref{fig:fig6}).
Meanwhile, GAN-based generative methods can improve semantic consistency to some extent. However, constrained by the inherent training instability and limited controllability of GANs, these methods often generate reconstructed images with noticeable blurring, artifacts, or local structural distortions, resulting in degraded perceptual quality, as shown in Fig.\,\ref{fig:fig8}. Semantic consistency loses much of its practical value if the reconstructed image suffers from severe perceptual degradation.
Although PICS achieves impressive perceptual quality scores (Fig.\,\ref{fig:fig8}) at ultra-low bitrates, this comes at the cost of compromised semantic consistency, as shown in Fig.\,\ref{fig:fig6}. Specifically, it relies heavily on generative priors, which grants the model excessive freedom, resulting in significantly poorer semantic alignment.
PerCo underperforms SDGIC in both semantic consistency (Fig.\,\ref{fig:fig6}) and perceptual quality (Fig.\,\ref{fig:fig8}) due to the relative sparsity of its guidance information. DDCM also benefits from diffusion priors, but its iterative reconstruction process usually introduces a high computational cost, reducing its decoding efficiency and practical cost-effectiveness. Overall, SDGIC achieves superior comprehensive performance across all evaluated metrics. This advantage mainly comes from the proposed semantic-disambiguation-guided framework, where the text caption serves as a global semantic anchor, the HCI provides dense visual anchoring, and the RSRTs introduce reconstruction-relevant semantic constraints. Together, these complementary guidance streams constrain the diffusion-based reconstruction process and reduce semantic ambiguity during generation.

Fig.\,\ref{fig:fig7} further demonstrates that SDGIC achieves state-of-the-art performance on AFINE, indicating that the proposed method obtains the best overall reconstruction quality and validating its effectiveness. Furthermore, our method achieves highly competitive performance on traditional pixel-level fidelity metrics, as reported in Fig.\,\ref{fig:fig9}.

Fig.\,\ref{fig:fig10} presents a qualitative comparison at ultra-low bitrate. Our method achieves high semantic consistency with the original images while producing more realistic perceptual quality. Conventional codecs exhibit noticeable artifacts: JPEG suffers from severe blocking effects, while ELIC produces overly smoothed textures. In the first and third rows, generative approaches, including VQIR, PICS, and PerCo, fail to reconstruct text accurately, resulting in character distortion or semantic deviation. In contrast, our method preserves textual content and structural patterns with high consistency. In the second row, our method reconstructs the feathers and eye details more realistically and naturally than competing approaches, demonstrating superior visual coherence while maintaining high consistency in the parrot's eye and surrounding textures.

\vspace{-6pt}
\section{Discussion}
\label{sec:discussion}
\begin{figure}[pos=t]
    \centering
    \includegraphics[width=\linewidth]{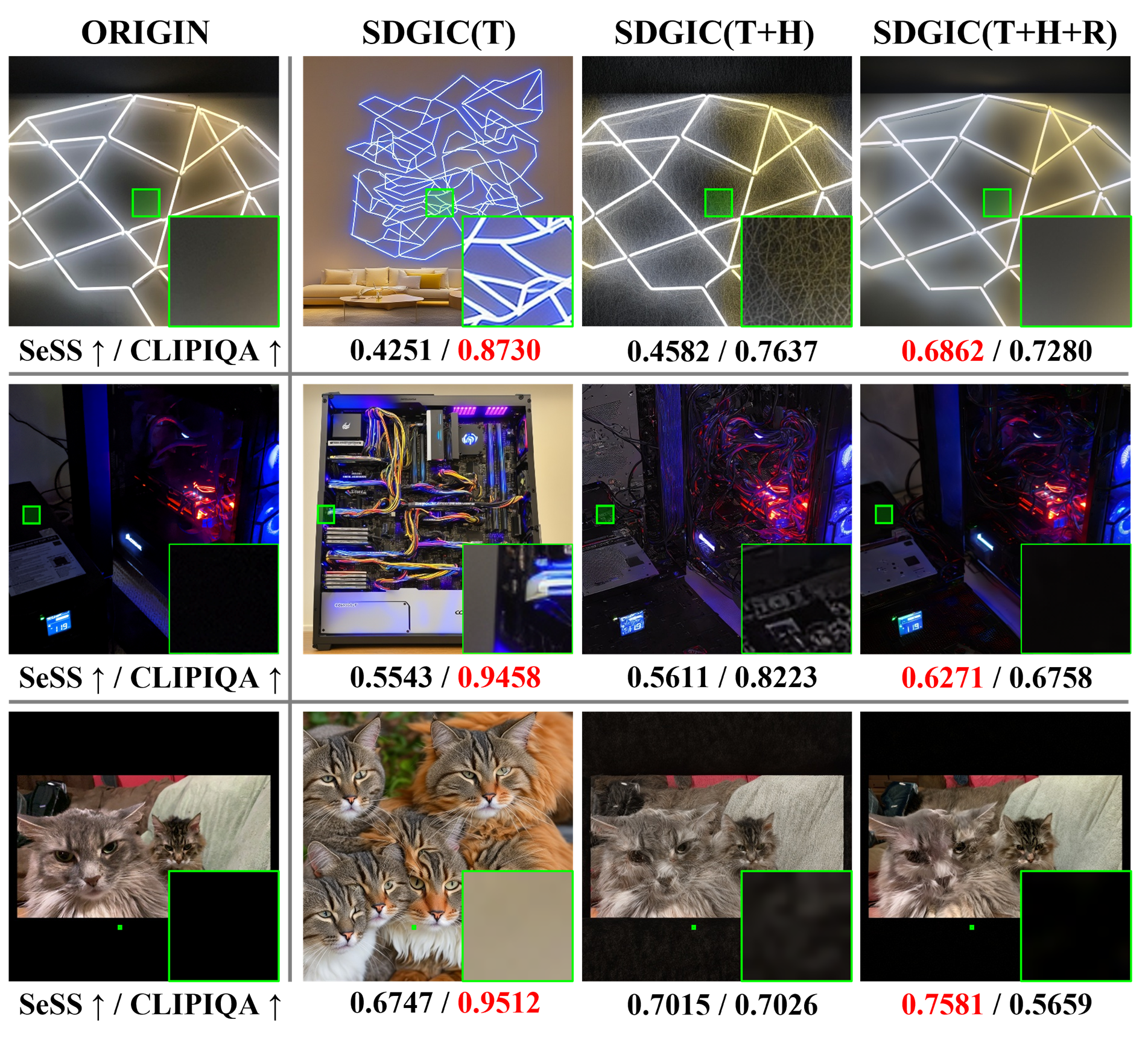} 
    \caption{Qualitative Comparison of Model Variants. SDGIC(T) achieves high perceptual scores at the expense of semantic consistency, while SDGIC(T+H+R) attains the optimal balance between both aspects.}
    \label{fig:fig11}
\end{figure}

\subsection{Ablation Analysis and Discussion}
\vspace{2pt}
\noindent\textbf{Effectiveness of Semantic Disambiguation Guidance.}
\begin{table*}[t]
    \vspace{2pt}
    \centering 
    \caption{Ablation analysis of semantic-disambiguation guidance streams on CLIC2020. The contributions of the text caption, HCI, and RSRTs are evaluated in terms of semantic consistency, perceptual quality, and pixel consistency.}
    \label{tab:ablation_guidance}
    \renewcommand{\arraystretch}{1.35}
    \setlength{\arrayrulewidth}{0.5pt}
    \setlength{\tabcolsep}{3.8pt}
    {\rmfamily\fontsize{9pt}{11pt}\selectfont
    \begin{tabular}{l|c|ccccc|cc|cc}
        \noalign{\hrule height 1.0pt}
        \multirow{2}{*}{\textbf{Methods}} 
        & \multirow{2}{*}{\textbf{bpp}} 
        & \multicolumn{5}{c|}{\textbf{Semantic Consistency}} 
        & \multicolumn{2}{c|}{\textbf{Perceptual Quality}} 
        & \multicolumn{2}{c}{\textbf{Pixel Consistency}} \\
        \cline{3-11}
        & 
        & ClipScore$\uparrow$ 
        & LPIPS$\downarrow$ 
        & DISTS$\downarrow$ 
        & SeSS$\uparrow$ 
        & AFINE$\downarrow$ 
        & CLIPIQA$\uparrow$ 
        & MANIQA$\uparrow$ 
        & PSNR$\uparrow$ 
        & MS-SSIM$\uparrow$ \\
        \hline
        SDGIC(T) & 0.002 & 0.7334 & 0.7192 & 0.3741 & 0.4952 & 37.7893 & \textcolor{red}{0.9256} & \textcolor{red}{0.5688} & 9.3687 & 0.1493 \\
        SDGIC(T+H) & 0.039 & 0.9183 & 0.2964 & 0.1927 & 0.6806 & 44.9688 & 0.8348 & 0.5289 & 20.4921 & 0.7492 \\
        \rowcolor{lightblue} 
        SDGIC(T+H+R) & 0.039 & \textcolor{red}{0.9439} & \textcolor{red}{0.2683} & \textcolor{red}{0.1766} & \textcolor{red}{0.7013} & \textcolor{red}{34.1903} & 0.8130 & 0.5571 & \textcolor{red}{22.8031} & \textcolor{red}{0.8084} \\
        \noalign{\hrule height 1.0pt}
    \end{tabular}
    }
    \vspace{2pt}
\end{table*}
To evaluate the effectiveness of the proposed semantic-disambiguation guidance streams, we conduct comprehensive ablation studies. We train three model variants that progressively integrate the guidance components: SDGIC(T), guided only by the text caption; SDGIC(T+H), guided by the text caption and HCI; and SDGIC(T+H+R), the full model guided by the text caption, HCI, and RSRTs. Across all variants, the text caption occupies a fixed bitrate of 0.002 bpp. For SDGIC(T+H), the $\lambda_{\mathrm{ELIC}}$ for generating the HCI is set to $3 \times 10^{-4}$. For SDGIC(T+H+R), $\lambda_{\mathrm{ELIC}}$ is set to $2 \times 10^{-4}$, and the number of RSRTs is set to 1. We quantitatively and qualitatively assess the performance of each variant, as shown in Table.\,\ref{tab:ablation_guidance} and Fig.\,\ref{fig:fig11}.

As shown in Table.\,\ref{tab:ablation_guidance}, although SDGIC(T) achieves significant bitrate savings, relying only on the text caption leaves the reconstruction severely underconstrained, leading to clear degradation in semantic consistency metrics such as LPIPS and DISTS. SDGIC(T+H) outperforms SDGIC(T) across most metrics, confirming that HCI provides dense visual anchoring by preserving low-level visual evidence such as structure, color, and layout, thereby mitigating distortion and semantic ambiguity during reconstruction. SDGIC(T+H+R) achieves the best overall performance. By incorporating RSRTs, SDGIC(T+H+R) further improves both semantic and pixel-level consistency over SDGIC(T+H). This improvement indicates that, although the text caption and HCI provide global semantic and dense visual constraints, respectively, some source-specific reconstruction cues remain ambiguous under these two conditions. RSRTs provide reconstruction-aware residual semantic constraints optimized toward the downstream denoising objective, helping the decoder select the source-consistent reconstruction from multiple plausible candidates.
The qualitative comparisons in Fig.\,\ref{fig:fig11} further validate this analysis. SDGIC(T) exhibits obvious semantic deviations because the text caption alone cannot sufficiently constrain the diffusion prior. SDGIC(T+H) improves the global structure and visual appearance by introducing HCI, but some local details may still be inconsistent with the source image. In contrast, SDGIC(T+H+R), augmented with RSRTs, effectively reduces such source-inconsistent details and achieves better semantic consistency with the original image.

\begin{table}[t] 
    \vspace{2pt}
    \centering
    \caption{Ablation study of text caption length for global semantic anchoring on CLIC2020. Red and blue denote the best and second-best results, respectively.}
    \label{tab:caption_length}
    \renewcommand{\arraystretch}{1.32}
    \setlength{\arrayrulewidth}{0.5pt}
    \setlength{\tabcolsep}{3.6pt}
    {\rmfamily\fontsize{9pt}{10.8pt}\selectfont
    \begin{tabular}{c|c|cccc}
        \noalign{\hrule height 1.0pt}
        \multirow{2}{*}{\makecell{\textbf{Length}\\{\normalfont\footnotesize (Text Caption)}}}
        & \multirow{2}{*}{\textbf{bpp}}
        & \multirow{2}{*}{\textbf{ClipScore$\uparrow$}}
        & \multirow{2}{*}{\textbf{LPIPS$\downarrow$}}
        & \multirow{2}{*}{\textbf{DISTS$\downarrow$}}
        & \multirow{2}{*}{\textbf{SeSS$\uparrow$}} \\
        & & & & & \\
        \hline
        0 words  & 0.037 & 0.9410 & 0.2810 & 0.1862 & 0.5509 \\
        \rowcolor{lightblue} 
        20 words & 0.039 & \textcolor{blue}{0.9439} & \textcolor{red}{0.2662} & 0.1766 & \textcolor{red}{0.7013} \\
        40 words & 0.041 & \textcolor{red}{0.9451} & \textcolor{blue}{0.2689} & \textcolor{blue}{0.1738} & \textcolor{blue}{0.7003} \\
        70 words & 0.044 & 0.9421 & 0.2865 & \textcolor{red}{0.1712} & 0.6949 \\
        \noalign{\hrule height 1.0pt}
    \end{tabular}
    }
    \vspace{2pt}
\end{table}
\begin{table}[t] 
    \vspace{2pt}
    \centering
    \caption{Ablation study of HCI as a dense visual anchor compared with sparse guidance representations on CLIC2020. Red and blue denote the best and second-best results, respectively.} 
    \label{tab:hci_vs_sparse_guidance}
    \renewcommand{\arraystretch}{1.4}
    \setlength{\arrayrulewidth}{0.5pt}
    \setlength{\tabcolsep}{2.6pt}
    {\rmfamily\fontsize{8.6pt}{10.3pt}\selectfont
    \begin{tabular}{c|c|cccc}
        \noalign{\hrule height 1.0pt}
        \textbf{Guidance} 
        & \textbf{bpp} 
        & \textbf{\makecell{ClipScore $\uparrow$}} 
        & \textbf{\makecell{LPIPS $\downarrow$}} 
        & \textbf{\makecell{DISTS $\downarrow$}} 
        & \textbf{\makecell{SeSS $\uparrow$}} \\
        \hline
        Text+Semantic & 0.0251 & 0.7513 & \textcolor{blue}{0.6390} & 0.3276 & 0.5510 \\
        Text+Sketch   & 0.0226 & \textcolor{blue}{0.7932} & 0.6858 & \textcolor{blue}{0.3052} & \textcolor{blue}{0.5778} \\
        \rowcolor{lightblue} 
        Text+HCI      & 0.0254 & \textcolor{red}{0.9096} & \textcolor{red}{0.3124} & \textcolor{red}{0.1967} & \textcolor{red}{0.6778} \\
        \noalign{\hrule height 1.0pt}
    \end{tabular}
    }
    \vspace{2pt}
\end{table}
Notably, we further investigate the relatively high perceptual scores of SDGIC(T) and SDGIC(T+H). As shown in Fig.\,\ref{fig:fig11}, SDGIC(T) lacks sufficient source-specific constraints, causing semantic drift despite producing visually coherent outputs. Similarly, SDGIC(T+H) can generate visually pleasing textures with the help of HCI, but the remaining semantic ambiguity may still allow the diffusion prior to synthesize details that are perceptually plausible yet inconsistent with the original image. These observations explain why perceptual metrics alone may be misleading under ultra-low bitrate settings. SDGIC(T+H+R) addresses this issue by introducing RSRTs, achieving a better balance between perceptual quality and semantic consistency. Furthermore, the ablation study on the text caption component in Table.\,\ref{tab:caption_length} validates its positive impact on semantic consistency and demonstrates its high bitrate efficiency, requiring only approximately 0.002 bpp. Overall, the three guidance streams in SDGIC play complementary roles in image reconstruction and work synergistically within the generative guidance process.

\vspace{2pt}
\noindent\textbf{Influence of Text Caption Length on Semantic Consistency.}
To validate the effectiveness of using a concise text caption as the global semantic anchor, we ablate the caption length within the SDGIC framework on CLIC2020. We compare four settings: no caption, and captions of 20, 40, and 70 words. As shown in Table.\,\ref{tab:caption_length}, adding a text caption consistently improves semantic consistency over the no-text baseline, confirming its role in providing global semantic constraints.
Notably, the 70-word caption does not outperform the 20-word and 40-word variants. This indicates that overly detailed captions may introduce semantic noise and redundancy due to the modality gap. In contrast, a concise caption provides robust high-level semantic priors, helping anchor the main objects and scene context while avoiding ambiguous or unnecessary details that may interfere with diffusion-based reconstruction.

\vspace{2pt}
\noindent\textbf{Effectiveness of HCI versus Sparse Guidance Representations.}
\begin{table}[t]
    \vspace{2pt}
    \centering
    \caption{Computational complexity and model parameters of different methods on CLIC2020. ET and DT denote encoding time and decoding time, respectively.}
    \label{tab:complexity_comparison}
    \renewcommand{\arraystretch}{1.28}
    \setlength{\arrayrulewidth}{0.5pt}
    \setlength{\tabcolsep}{5pt}
    {\rmfamily\fontsize{8pt}{9.6pt}\selectfont
    \begin{tabular}{cc|c|c|c|c}
        \noalign{\hrule height 1.0pt}
        \multicolumn{2}{c|}{\textbf{Methods}} 
        & \textbf{Steps} 
        & \textbf{ET(s)} 
        & \textbf{DT(s)} 
        & \textbf{Params} \\
        \hline
        
        \multirow{2}{*}{\shortstack{Hand-crafted}} 
          & BPG & / & 0.25 & 0.20 & / \\
          & VTM & / & 22.28 & 0.12 & / \\
        \hline
        
        \multirow{2}{*}{\shortstack{VAE-based}} 
          & ELIC & / & 0.37 & 0.22 & $4.62{\times}10^7$ \\
          & MLIC++ & / & 0.45 & 0.55 & $1.51{\times}10^8$ \\
        \hline
        
        \multirow{1}{*}{\shortstack{GAN-based}} 
          & HiFiC & / & 1.87 & 3.81 & $1.48{\times}10^8$ \\
        \hline
        
        \multirow{10}{*}{\shortstack{Diffusion-based}} 
          & \multirow{3}{*}{PICS} & 10 & 210.99 & 26.43 & $3.61{\times}10^9$ \\
          & & 30 & 210.99 & 41.98 & $3.61{\times}10^9$ \\
          & & 50 & 210.99 & 57.64 & $3.61{\times}10^9$ \\
        \cline{2-6}
          & \multirow{3}{*}{PerCo} & 10 & 0.63 & 4.61 & $4.99{\times}10^9$ \\
          & & 30 & 0.63 & 13.43 & $4.99{\times}10^9$ \\
          & & 50 & 0.63 & 22.24 & $4.99{\times}10^9$ \\
        \cline{2-6}
          & DDCM & 1000 & 31.16 & 30.84 & $1.24{\times}10^9$ \\
        \cline{2-6}
          \rowcolor{lightblue}
          \cellcolor{white}
          & & 10 & 0.88 & 1.55 & $6.07{\times}10^9$ \\
          \rowcolor{lightblue}
          \cellcolor{white}
          & \multirow{-2}{*}{SDGIC} & 30 & 0.88 & 3.74 & $6.07{\times}10^9$ \\
          \rowcolor{lightblue}
          \cellcolor{white}
          & \multirow{-2}{*}{(Ours)} & 50 & 0.88 & 6.94 & $6.07{\times}10^9$ \\
        \noalign{\hrule height 1.0pt}
    \end{tabular}
    }
    \vspace{2pt}
\end{table}
To validate the effectiveness of HCI as a dense visual anchor, we compare it with common sparse guidance representations on CLIC2020. We train three image generation models from scratch, each conditioned on text paired with a different visual representation: Text+Semantic map, Text+Sketch, and Text+HCI. As shown in Table,\ref{tab:hci_vs_sparse_guidance}, Text+HCI outperforms the other two settings across most metrics. For example, it achieves an LPIPS score of 0.1967, reducing LPIPS by approximately 39.9\% and 35.5\% compared with Text+Semantic map and Text+Sketch, respectively. These results show that HCI preserves denser visual evidence through rate-distortion optimization, providing stronger constraints for diffusion-based reconstruction and reducing semantic ambiguity.

\subsection{Influence of Denoising Steps}
\vspace{2pt}
\begin{figure*}[t]
    \centering
    \includegraphics[width=1\textwidth]{Fig/Fig12.pdf} 
    \caption{Impact of Denoising Steps on Semantic Consistency. A quantitative ablation on CLIC2020 shows that increasing denoising steps consistently improves the semantic consistency (ClipScore, LPIPS, DISTS, SeSS) of reconstructed images.}
    \label{fig:fig12}
\end{figure*}
\begin{table*}[t]
    \vspace{2pt}
    \centering
    \caption{Performance impact of different image-to-text models used for global semantic anchoring. EP and DP denote encoder parameters and decoder parameters, respectively.}
    \label{tab:captioner_variants}
    \renewcommand{\arraystretch}{1.32} 
    \setlength{\arrayrulewidth}{0.5pt} 
    \setlength{\tabcolsep}{4pt}
    {\rmfamily\fontsize{9pt}{10.8pt}\selectfont
    \begin{tabular}{l|c|cccc|cc}
        \noalign{\hrule height 1.0pt}
        \multicolumn{1}{c|}{\textbf{Model}} 
        & \textbf{bpp} 
        & \textbf{ClipScore$\uparrow$} 
        & \textbf{SeSS$\uparrow$} 
        & \textbf{AFINE$\downarrow$} 
        & \textbf{CLIPIQA$\uparrow$} 
        & \textbf{EP} 
        & \textbf{DP} \\
        \hline
        SDGIC (GIT)    & 0.0389 & 0.9424 & 0.6793 & 35.54 & 0.8031 & $9.13{\times}10^8$ & $1.58{\times}10^9$ \\
        SDGIC (BLIP-2) & 0.0395 & 0.9434 & 0.7002 & 34.94 & 0.8132 & $4.48{\times}10^9$ & $1.58{\times}10^9$ \\
        SDGIC (LMM)    & 0.0396 & 0.9440 & 0.7013 & 34.89 & 0.8130 & /                  & $1.58{\times}10^9$ \\
        \noalign{\hrule height 1.0pt}
    \end{tabular}
    }
    \vspace{2pt}
\end{table*}
We further investigate the impact of denoising steps on semantic consistency. As shown in Fig.\,\ref{fig:fig12}, we vary the number of denoising steps from 10 to 50. The results show that increasing denoising steps consistently improves semantic consistency. However, this improvement may come at the cost of higher decoding complexity, as evidenced by the results in Table.\,\ref{tab:complexity_comparison}.

\subsection{Complexity and Efficiency Analysis}
\vspace{2pt}
\noindent\textbf{Quantitative Comparison.}
Table.\,\ref{tab:complexity_comparison} provides a complexity analysis in terms of encoding time (ET), decoding time (DT), and model parameters (Params). We compare SDGIC with diverse baselines, including traditional codecs (BPG\cite{bellard2015bpg}, VTM\cite{bross2021overview}), VAE-based models (ELIC\cite{he2022elic}, MLIC++\cite{jiang2025mlic++}), GAN-based methods (HiFiC\cite{mentzer2020high}), and diffusion-based approaches (PICS\cite{lei2023text+}, PerCo\cite{careil2023towards}, DDCM\cite{ohayon2025compressed}). For a fair comparison, we use the same BLIP-2\cite{li2023blip} model as PerCo for text caption generation.

 Non-generative compression schemes such as BPG, VTM, ELIC, and MLIC++ achieve faster inference due to their deterministic and non-iterative decoding processes, as shown in Table.\,\ref{tab:complexity_comparison}. Among diffusion-based methods, PICS exhibits the highest latency because of its complex iterative optimization process. Similarly, DDCM incurs substantial time costs, as its architecture typically requires approximately 1,000 iterative steps for encoding and decoding. Although PerCo improves upon PICS and DDCM, SDGIC achieves a better overall balance. Specifically, SDGIC has competitive encoding time and achieves the fastest decoding speed among the compared generative methods.
In terms of model parameters, all diffusion-based methods, including PICS, PerCo, DDCM, and SDGIC, remain within the same order of magnitude. However, their parameter counts are higher than those of VAE-based and GAN-based methods, indicating greater VRAM requirements. This is an inherent limitation of diffusion models and will be an important direction for our future work.

\vspace{2pt}
\noindent\textbf{Flexible Model Deployment.}
Specifically, we evaluate three captioners with different model scales, including the lightweight GIT\cite{wang2022git}, the medium-scale BLIP-2, and Gemini 2.5 Pro, and assess their impact on reconstruction performance using the CLIC2020 dataset. Table.\,\ref{tab:captioner_variants} reports the reconstruction performance, encoder parameters (EP), and decoder parameters (DP).
The results show that performance variations caused by different captioners remain within an acceptable range without obvious quality degradation. This stability mainly comes from the fact that SDGIC uses concise captions as global semantic anchors, rather than relying on text to describe pixel-level details. Therefore, the framework is robust to different captioners. Users can flexibly select the captioner according to deployment requirements and available hardware resources.

\section{Conclusion}
\label{sec:conclusion}
In this work, we propose a semantic-disambiguation-guided generative image compression framework (SDGIC) for ultra-low-bitrate image compression. Specifically, a concise text caption provides global semantic anchoring, an HCI supplies dense visual evidence, and RSRTs serve as reconstruction-aware residual semantic constraints optimized through the downstream denoising objective. Meanwhile, the proposed DPCD integrates these guidance signals through conditioning routes matched to their respective information roles, enabling controllable image reconstruction. Experimental results demonstrate that SDGIC achieves superior overall reconstruction performance. This work provides a practical and interpretable approach for reducing source-inconsistent semantic deviations in generative image compression, and offers useful insights into role-aware guidance design for reliable ultra-low-bitrate visual communication.

\bibliographystyle{elsarticle-num}
\bibliography{cas-refs}

\end{document}